\documentclass[conference]{ieeeconf}
\IEEEoverridecommandlockouts

\usepackage[style=ieee,hyperref,natbib=true,backend=bibtex,firstinits,doi=false,%
     mincitenames=1,maxcitenames=2,maxbibnames=99,sorting=none,terseinits=false,hyperref=true]{biblatex}

\bibliography{references.bib}

\usepackage{amsmath,amssymb,amsfonts}
\usepackage{subfig}
\renewbibmacro*{bbx:savehash}{}
\defbibheading{bibliography}[\bibname]{\section*{References}}

\usepackage{url}
\usepackage{graphicx}
\usepackage[usenames,dvipsnames]{xcolor}
\usepackage[draft]{fixme}
\fxsetup{theme=color}

\usepackage{amsmath}
\usepackage{amssymb}
\usepackage{textcomp}
\usepackage{siunitx}
\usepackage{acronym}
\usepackage{booktabs}
\usepackage{threeparttable}
\usepackage{multirow}

\usepackage{calc}

\usepackage[capitalize]{cleveref}

\usepackage{tikz}
\usetikzlibrary{arrows}
\usetikzlibrary{positioning,calc}
\usetikzlibrary{decorations.pathreplacing}
\usetikzlibrary{decorations.markings}
\usetikzlibrary{fit}
\usetikzlibrary{shapes.callouts}
\usetikzlibrary{shapes.geometric}
\usetikzlibrary{matrix}

\usepackage{pgfplots}
\pgfplotsset{compat=1.9}
\usepgfplotslibrary{groupplots}
\usepgfplotslibrary{units}
\usepgfplotslibrary{statistics}
\usepgfplotslibrary{colorbrewer}
\usepackage{pgfplotstable}

\tikzdeclarecoordinatesystem{rel}{%
	\tikzset{cs/.cd,x=0pt,y=0pt,#1}%
	\pgfpointlineattime{(\tikz@cs@x)}%
	{\pgfpointanchor{\tikz@pp@name{\tikz@cs@node}}{south west}}%
	{\pgfpointanchor{\tikz@pp@name{\tikz@cs@node}}{south east}}%
	\edef\tikz@cs@x{\the\pgf@x}%
	\pgfpointlineattime{(\tikz@cs@y)}%
	{\pgfpointanchor{\tikz@pp@name{\tikz@cs@node}}{south west}}%
	{\pgfpointanchor{\tikz@pp@name{\tikz@cs@node}}{north west}}%
	\pgfpoint{\tikz@cs@x}{\pgf@y}%
}

\tikzdeclarecoordinatesystem{rel2}{%
	\tikzset{cs/.cd,x=0pt,y=0pt,#1}%
	\pgfpointlineattime{(\tikz@cs@x / 100)}%
	{\pgfpointanchor{\tikz@pp@name{\tikz@cs@node}}{south west}}%
	{\pgfpointanchor{\tikz@pp@name{\tikz@cs@node}}{south east}}%
	\edef\tikz@cs@x{\the\pgf@x}%
	\pgfpointlineattime{(\tikz@cs@y / 100)}%
	{\pgfpointanchor{\tikz@pp@name{\tikz@cs@node}}{south west}}%
	{\pgfpointanchor{\tikz@pp@name{\tikz@cs@node}}{north west}}%
	\pgfpoint{\tikz@cs@x}{\pgf@y}%
}

\newcommand\currentcoordinate{\the\tikz@lastxsaved,\the\tikz@lastysaved}

\pgfplotsset{
    boxplot discard if not/.style 2 args={
        /pgfplots/boxplot/data filter/.code={
            \edef\tempa{\thisrow{#1}}
            \edef\tempb{#2}
            \ifx\tempa\tempb
            \else
                
            \fi
        }
    }
}
\pgfplotsset{
    discard if not/.style 2 args={
        x filter/.code={
            \edef\tempa{\thisrow{#1}}
            \edef\tempb{#2}
            \ifx\tempa\tempb
            \else
                
            \fi
        }
    }
}
\pgfplotsset{filter discard warning=false}
\makeatletter
\pgfplotsset{
    boxplot/hide outliers/.code={
        \def\pgfplotsplothandlerboxplot@outlier{}%
    }
}

\usepackage{letltxmacro,xcolor,xparse}
\usepackage[export]{adjustbox}

\usepackage{ifthen}
\IfFileExists{graphicscache.sty}{%
  \usepackage[dpi=440]{graphicscache}%
}{%
}

\usepackage{dblfloatfix}

\tikzset{
  font=\sffamily\footnotesize,
  m/.style={draw, rounded corners, fill=yellow!20, align=center}
}

\pgfplotsset{
        compat=1.7,
    }

\acrodef{VR}{Virtual Reality}
\acrodef{IMU}{Inertial Measurement Unit}
\acrodef{DoF}{Degree of Freedom}

\definecolor{t5}{RGB}{235, 172, 35}
\definecolor{t8}{RGB}{184, 0, 88}

\DeclareRobustCommand{\taskb}[1]{\strut\tikz{\node[fill=#1,rounded corners=1pt,minimum height=1.5ex,minimum width=1.5ex]{};}}

\def\BibTeX{{\rm B\kern-.05em{\sc i\kern-.025em b}\kern-.08em
    T\kern-.1667em\lower.7ex\hbox{E}\kern-.125emX}}
\begin{document}

\title{\LARGE \bf
Self-Centering 3-DoF Feet Controller for Hands-Free Locomotion Control in Telepresence and Virtual Reality
}

\author{Raphael Memmesheimer, Christian Lenz, Max Schwarz,\\ Michael Schreiber, and Sven Behnke
\thanks{All authors are with the Autonomous Intelligent Systems group, Computer Science Institute VI – Intelligent Systems and Robotics, Lamarr
Institute for Machine Learning and Artificial Intelligence, and Center for
Robotics, University of Bonn; {\tt memmesheimer@ais.uni-bonn.de}}%
}

\maketitle

\begin{abstract}

We present a novel seated feet controller for handling 3 \ac{DoF} aimed to control locomotion for telepresence robotics and virtual reality environments.
Tilting the feet on two axes yields in forward, backward and sideways motion. In addition, a separate rotary joint allows for rotation around the vertical axis. Attached springs on all joints self-center the controller. The HTC Vive tracker is used to translate the trackers' orientation into locomotion commands.
The proposed self-centering feet controller was used successfully for the ANA Avatar XPRIZE competition, where a naive operator traversed the robot through a longer distance, surpassing obstacles while solving various interaction and manipulation tasks in between.
We publicly provide the models of the mostly 3D-printed feet controller for reproduction.
\end{abstract}

\section{Introduction}

\begin{figure}
    \centering \vspace*{-3ex}
    \includegraphics[width=.7\linewidth]{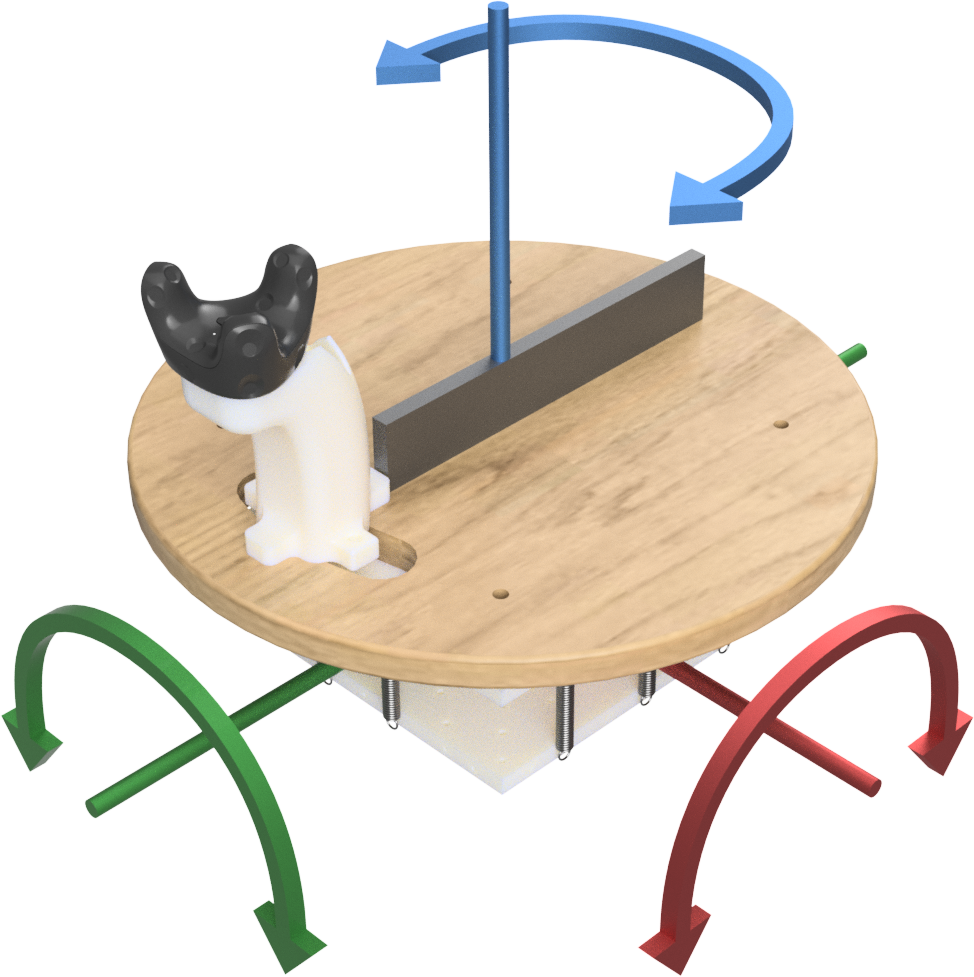}\vspace*{-2ex}
    \caption{Feet controller with its controllable axes.\vspace*{-3ex}}
    \label{fig:overview}
\end{figure}

Controller alternatives for virtual reality environments or telepresence robotics, especially in situations where the hands are already reserved for other controlling schemes, are underrepresented.
Even so, there are various controlling alternatives like human brain interfaces, 3D rudders, treadmills \cite{lichtenstein2007feedback}, tongue controllers \cite{DBLP:journals/tbe/Struijk06,huo2008magneto}, eye tracking \cite{shirakura2005development}, they all have their individual advantages and disadvantages.
Seated locomotion controllers are relaxing for operators and allow long-term use. They further can be useful for people with disabilities.
In this paper, we present a novel 3-\ac{DoF} locomotion controller that utilizes the operator's feet for control and is suitable for seated operation. A rendering of the proposed locomotion controller and the controllable axes is shown in \cref{fig:overview}. The locomotion controller was designed for our immersive and mobile teleoperation robot in context of the ANA Avatar XPRIZE competition\footnote{\url{https://avatar.XPRIZE.org/prizes/avatar}} with the following requirements:

\noindent \textit{Intuitive Design}: The controller should be designed with an emphasis on user-friendliness, requiring minimal training overhead without prior experience.

\noindent \textit{Hands-Free Operation}: In our system, the operator is coupled with an exoskeleton for arms, a \ac{VR} glove and a \ac{VR} headset, therefore only hands-free operation to ensure intuitive 3-\ac{DoF} locomotion is possible.

\noindent \textit{Endurance}: The designed competition track includes solving multiple tasks, therefore must be capable of long-term operation without causing user exhaustion.

\noindent \textit{Control Accuracy}: The controller shall exhibit high control accuracy, enabling precise and fine-grained locomotion maneuvers. 

\noindent \textit{Calibration}: The design should minimize calibration overhead; therefore HTC Vice trackers were utilized and only a single state calibration is required.

\noindent  \textit{State Re-initialization}: The controller must provide automatic state re-initialization capabilities, when the user removes its feet from the controller. This is solved by installing a self-centering spring mechanism for translational and rotational axes. 

\noindent \textit{Configurability}: The controller shall allow for configurability of the pressure settings. Users should have the ability to adjust pressure parameters to suit their preferences and specific application needs, enhancing the overall versatility of the device.

By adhering to these design requirements, the locomotion controller will offer an operator-friendly and efficient solution for various locomotion control applications.
The proposed feet controller was successfully used for our participation in the ANA Avatar XPRIZE competition \cite{lenz2023nimbro}.
The 3D models to reproduce the proposed controller are made publicly available on Printables\footnote{\url{https://www.printables.com/model/961854}} and GitHub\footnote{\url{https://github.com/AIS-Bonn/hands-free_locomotion_controller}}. 

\section{Related Work}

An extensive summary of various hands-free locomotion controllers for avatar robots has been assembled by \citet{wittendorp2020locomotion}.
Various concepts focusing on body-leaning and feet-controllers have been drafted. Four concepts have then been realized and evaluated in a \ac{VR} environment with a focus on learnability.
They conclude that body-leaning controllers are more intuitive for operators.
We argue that intuitive feet-based locomotion control still can be  achieved by employing a controller that is self-centering and employs configurable tension.
Further, the operator should not be confronted with the calibration overhead, and the calibration should be minimal.
\citet{DBLP:conf/uist/SaraijiSKMI18} presented a feet controller for controlling attached robotic arms to a human. Their controller is based on a sock with a flexion band and a HTC Vive tracker for position estimates.
For \ac{VR} locomotion, the point and teleport technique has been well established \cite{bozgeyikli2016point}. While being applicable in \ac{VR} in telepresence, such a technique would bring overhead like the robots, requiring additional sensors to map the environment and autonomous navigation capabilities.

\citet{otaran2021walking} present an ankle-actuated seated VR locomotion interface. Sensorized active joints allow linear locomotion generation by tapping the feet on a platform of the device to estimate the footsteps.
As the device is limited to 1-\ac{DoF} walking such that the steering is achieved by an additional VR headset. Similar to our proposed locomotion controller, their locomotion interface is operated in a relaxing seated position.
Interesting to note is that the locomotion interface can display haptics in the form of varying ground textures using the motorized joint.
The steps are estimated by a simple threshing algorithm around the platform rotation.
\citet{carmichael2020spring} presented Spring Stepper, a seated locomotion device that registers naturally walking steps to map those to \ac{VR} movement.
\citet{darken1997omni} presented an omnidirectional treadmill which allows the user to walk and jog in any direction of movement. The overall treadmill is a combination of two perpendicular treadmills on two layers.
\citet{von2020podoportation} presented a foot-based locomotion controller based on 3D position of the user’s feet and the pressure applied to the sole as input modalities.

In the ANA Avatar XPRIZE competition, various locomotion controllers have been employed. Among them were game pads\cite{vaz2022immersive}, space mouses \cite{haruna2023avatar}, handheld \ac{VR} controllers \cite{lentini2019alter}, omnidirectional \ac{VR} treadmills \cite{cakmak2014cyberith, dafarra2022icub3}, feet pedals \cite{lee2023foot, correia2024immersive} and 3D rudders \cite{van2022comes}.


\section{3-\ac{DoF} Locomotion Controller}

\begin{figure}
  \input{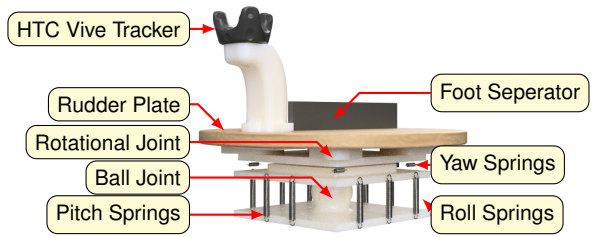}
  \caption{Self-centering 3D rudder design with individually tunable springs for intuitive locomotion control.}
  \vspace{-1ex}
  \label{fig:rudder}
\end{figure}

\begin{figure*}
  \centering
  \subfloat[Side]{\includegraphics[width=0.2\textwidth]{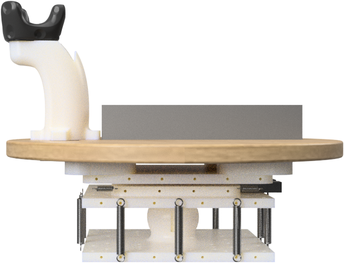}}\hspace{1cm}
  \subfloat[Home]{\includegraphics[width=0.2\textwidth]{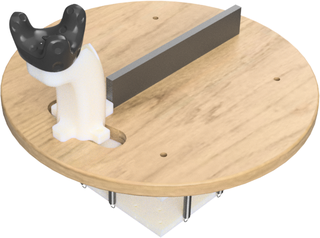}}\hspace{1cm}
  \subfloat[Top]{\includegraphics[width=0.2\textwidth]{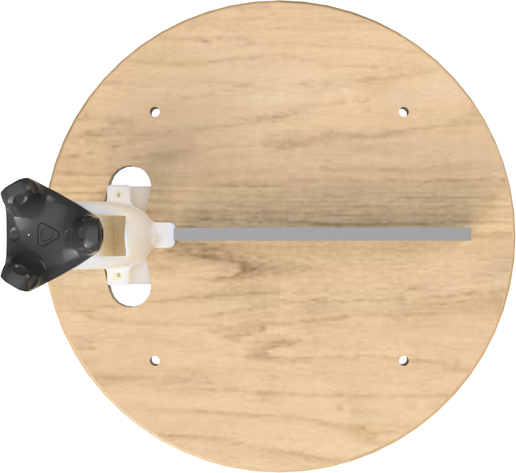}}
  \caption{Rendered views from for the proposed locomotion controller.}
  \label{fig:views}
\end{figure*}

The proposed feet controller draws inspiration from the 3DRudder, yet additionally simplifies the use of telepresence control by being self-centering and facilitating simplified calibration.
A detailed annotation of the proposed controller is given in \cref{fig:rudder} with additional views of the locomotion controller being depicted in \cref{fig:views}.
\begin{figure}
  \centering
  \subfloat[Forward]{\includegraphics[width=0.3\linewidth]{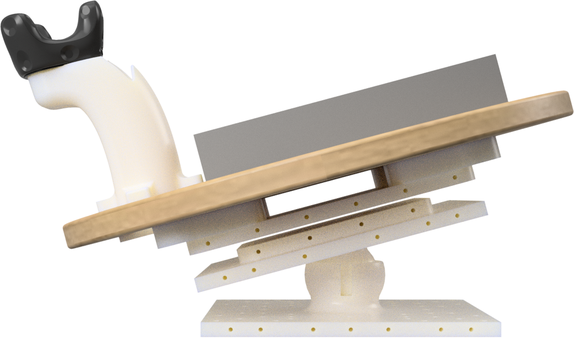}}\quad
  \subfloat[Backward]{\includegraphics[width=0.25\linewidth]{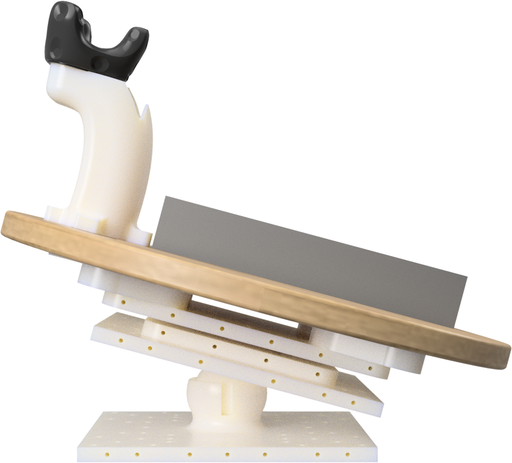}}\\
  \subfloat[Turning Left]{\includegraphics[width=0.3\linewidth]{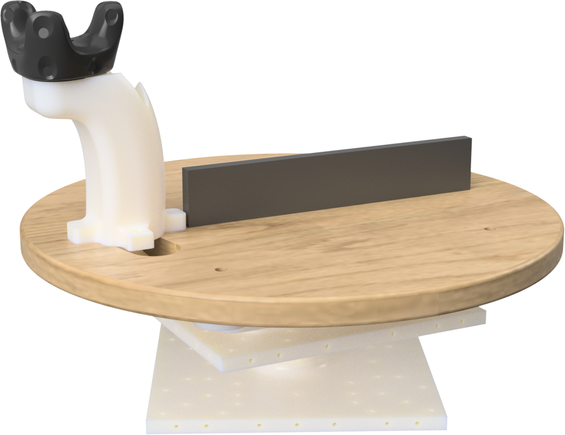}} \quad
  \subfloat[Turning Right]{\includegraphics[width=0.3\linewidth]{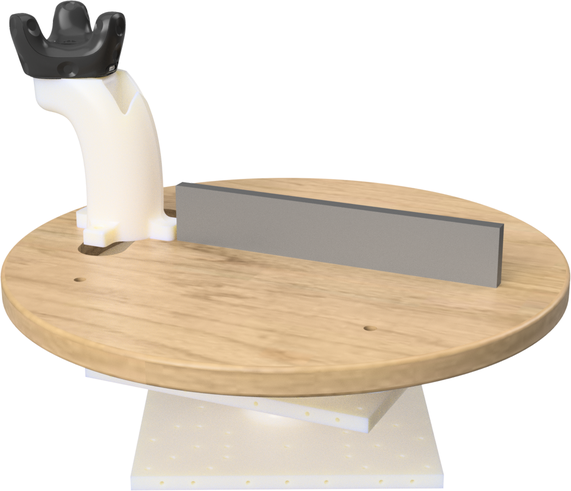}} \\
  \subfloat[Left]{\includegraphics[width=0.3\linewidth]{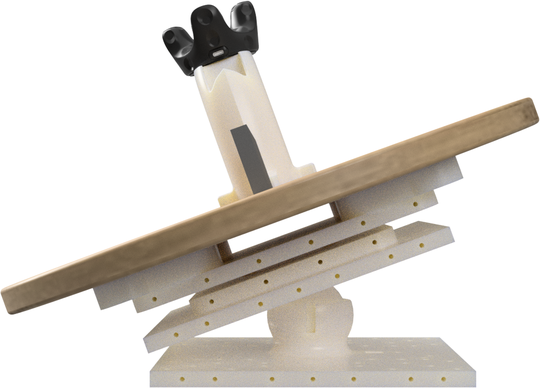}}\quad
  \subfloat[Right]{\includegraphics[width=0.3\linewidth]{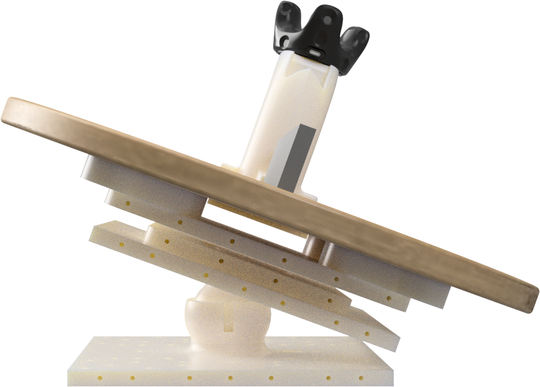}}\\
  \caption{Controller configurations and their resulting locomotion commands (simplified without springs).}
  \label{fig:motions}
\end{figure}

\begin{figure}
    \centering
    \includegraphics[width=\linewidth]{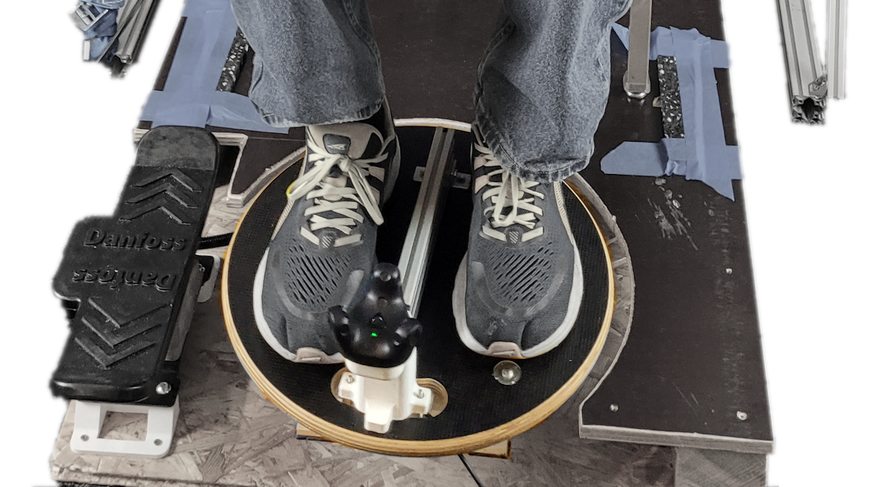}
    \caption{Locomotion controller in operation.}
    \label{fig:locomotion_controller_in_operation}
\end{figure}
\cref{fig:locomotion_controller_in_operation} shows the controller in operation of our immersive telepresence system \cite{lenz2023nimbro, schwarz2023robust}.  


The controller consists of a ball joint that allows the controller to tilt left, right, forward and backward. A separate rotary joint, utilizing a 50$\times$70$\times$14\,mm thrust bearing, provides flexible control over the rotary resistance. 
Roll and pitch components were mapped to the linear velocities, and the yaw component was mapped to the angular velocity. Linear and rotational velocities are individually scaled.
See \cref{fig:motions} for visualization of the locomotion and their respective rudder state. 
The initial state is assumed to be the origin, which is further simplified by the self-centering property.
The self-centering mechanism ensures that no movement is recorded if the feet are not positioned on the rudder surface. 
The self-centering mechanism is implemented through the installation of 10 tension springs, comprising three on the left and right, two on the front and back, each with a diameter of $6.3$\,mm and a length of $22.5$\,mm. 
The difference in the number of springs was chosen to reflect intuitive muscular forces for back- and forward movement, as well as sideways movement.
The rotary axis employs four tension springs with diameters of $5.5$\,mm and a length of $20.2$\,mm each. Resistance on all axes can be configured by the number of installed springs and their characteristics.
The rotary axis is limited mechanically to prevent the rotary springs from overextending. The ball joint also limits mechanically the pitch and roll range.

To prevent unintended motion, an inactive zone on all axes is defined before translating the pose information to locomotion commands.
The controller is supposed to be fixed on a static planar surface and therefore results in 3-\ac{DoF} control.

The feet surface was realized using a rounded wooden furniture roller. To facilitate the blind repositioning of the feet, a central separator has been installed. To enhance visibility in the base stations, the HTC Vive 3.0 tracker is positioned in front above the controller surface.

The rudder was developed through multiple iterations and experiments. Initially, the design was based on an \ac{IMU}, but was found to only be suitable for moving sideways, forward, and backwards as the rotational component tended to drift too much. Furthermore, we developed an iteration, where the ball-joint was replaced by a force-torque sensor. We observed that this design iteration was unintuitive for the user due to its excessive rigidity. Additional experiments on the force-torque-based version, such as the incorporation of dampers to address the rigidity issue, resulted in imprecise control commands.





\section{Experiments}

\pgfplotstableread[col sep=comma]{data/task_timings.csv}{\loadedtable}
\pgfplotstabletranspose[colnames from=Team]\transposedtable\loadedtable
\begin{figure}
 \centering%
 \resizebox{\columnwidth}{!}{%
 \begin{tikzpicture}[
    ppin/.style={pin edge={latex-,red,thick},pin distance=8pt,font=\sffamily\scriptsize,draw=red,rounded corners},
  ]
  \begin{axis}[
      anchor=north,
      height=7cm,
      clip=false,
      xmin=0,
      xmax=6,
      width=.7\columnwidth,
      nodes near coords xbar stacked configuration/.style={},
      xbar stacked,
      bar width=4pt,
      axis x line*=left,
      axis y line*=left,
      xtick align=outside,
      xtick pos=bottom,
      xtick distance=2,
      minor x tick num=3,
      ytick=data,
      yticklabels from table={\transposedtable}{colnames},
      ytick pos=left,
      y dir=reverse,
      xlabel={\footnotesize Time [min:sec]},
      x filter/.code={\pgfmathparse{#1/60}},
      xticklabel={ 
        \pgfmathsetmacro\hours{floor(\tick)}%
        \pgfmathsetmacro\minutes{(\tick-\hours)*0.6}%
        \pgfmathprintnumber{\hours}:\pgfmathprintnumber[fixed, fixed zerofill, skip 0.=true, dec sep={}]{\minutes}%
        },
      printtime/.style={%
        point meta=x,
        nodes near coords style={black},
        nodes near coords={
            \pgfkeys{/pgf/fpu=true}
            \pgfmathparse{\pgfplotspointmeta}
            \pgfmathsetmacro\hours{floor(\pgfplotspointmeta)}%
            \pgfmathsetmacro\minutes{(\pgfplotspointmeta-\hours)*0.6}%
            \pgfmathprintnumber{\hours}:\pgfmathprintnumber[fixed, fixed zerofill, skip 0.=true, dec sep={}]{\minutes}%
            \pgfkeys{/pgf/fpu=false}
        }
      },
      cycle list={
        {draw=none,fill=t5},
        {draw=none,fill=t8},
      },
      every plot/.style={draw=none}
    ]


    \addplot table [x index=1, y expr=\coordindex] {\transposedtable};
   \addplot+[printtime] table [x index=2, y expr=\coordindex] {\transposedtable};
  \end{axis}
 \end{tikzpicture} 
 }%
 \vspace{-0.2cm}
\caption{ANA Avatar XPRIZE finals task completion time for \taskb{t5}\,Task~5 (approx. 40\,m of locomotion) and \taskb{t8}\,Task~8 (navigate around obstacles), including all trials in which both tasks were solved.}
 \label{fig:locomotion_stats}
\end{figure}
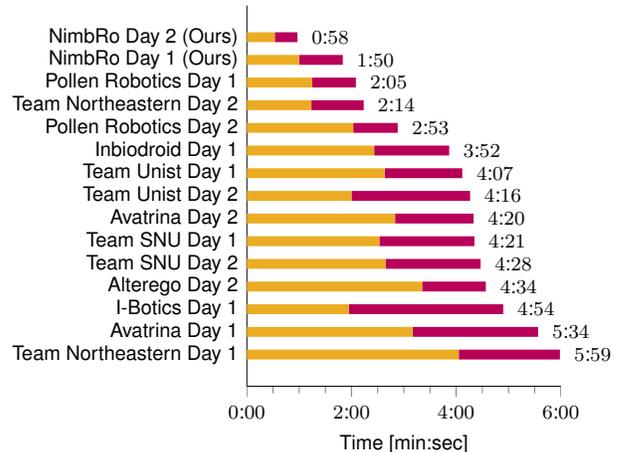

The proposed locomotion controller has been evaluated in the context of the ANA Avatar XPRIZE competition~\cite{behnke202310}. The described experiments are two-fold. First, we analyze the two locomotion-centric tasks (out of a total of ten tasks):
Task~5 involving traversing a distance of approximately $40$\,m and Task~8 where the operator had to traverse the robot through a field of obstacles. We extracted the tasks timings using the official live-stream footage for all teams on both competition days.
\cref{fig:locomotion_stats} shows the timings for all trials, successfully solving both tasks of the ANA Avatar XPRIZE competition.
During both competition days, our system demonstrated the fastest locomotion timing among all participating teams. 
It is noteworthy that our system performed faster on both tasks of the Day~2 run together than all other systems on the first task, highlighting the performance of the controller and the system.
Naturally, these timings reflect a combination of system components, such as the mobile platform and the locomotion controller. Therefore, they can be only interpreted as whole system design decisions (see \cite{lenz2023nimbro} for a complete system description and evaluation). As depicted in the figure, the system proved to be suitable for intuitively traversing obstacles. We increased the platform acceleration and maximum speed parameters between Day~1 and Day~2, demonstrating that the locomotion controller remains intuitive throughout different operation speeds.
\pgfplotstableread[col sep=comma]{data/rudder_day2.csv}{\datatable}
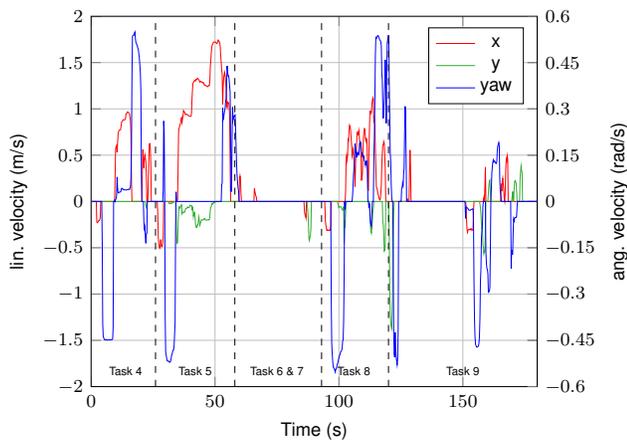
\begin{figure}
    \centering
    \resizebox{\columnwidth}{!}{%
    \begin{tikzpicture}
        \begin{axis}[
            xlabel={Time (s)},
            ylabel={lin. velocity (m/s)},
            grid=both,
            ytick distance=0.5,
            ymin=-2.0, ymax=2.0,
            xmin=0, xmax=180,
            major grid style={line width=.2pt,draw=gray!50},
            minor grid style={line width=.1pt,draw=gray!50},
        ]
        \addplot [
            mark=none,
            color=red,
        ] table [x=Time, y=x_vel] {\datatable};
        \label{plot_one}
        \addplot [
            mark=none,
            color=green!60!black!80,
        ] table [x=Time, y=y_vel] {\datatable};
        \label{plot_two}
        \end{axis}

        \begin{axis}[
            ylabel={ang. velocity (rad/s)},
            axis y line*=right, 
            axis x line=none,
            legend pos=south west,
            grid=none,
            ymin=-0.6, ymax=0.6,
            xmin=0, xmax=180,
            ytick distance=0.15,
            major grid style={line width=.2pt,draw=gray!50},
            minor grid style={line width=.1pt,draw=gray!50},
            yticklabel pos=right, 
            legend pos=north east,
        ]
        \addlegendimage{/pgfplots/refstyle=plot_one}\addlegendentry{x}
        \addlegendimage{/pgfplots/refstyle=plot_two}\addlegendentry{y}
        \addplot [
            mark=none, 
            color=blue,
        ] table [x=Time, y=yaw] {\datatable};
        \addlegendentry{yaw}
        \draw[dashed, black](26,0)--(26,1500);
        \draw[dashed, black](58,0)--(58,1500);
        \draw[dashed, black](93,0)--(93,1500);
        \draw[dashed, black](120,0)--(120,1500);

        \node[align=center] at (14,50) {\tiny Task 4};
        \node[align=center] at (42,50) {\tiny Task 5};
        \node[align=center] at (75,50) {\tiny Task 6 \& 7};
        \node[align=center] at (106,50) {\tiny Task 8};
        \node[align=center] at (150,50) {\tiny Task 9};
        \end{axis}
        
    \end{tikzpicture}
    }%
    \vspace{-0.2cm}
    \caption{Linear and angular velocities of NimbRo Day~2 run for Tasks~4 through 9.}
    \label{fig:velocity_time}
\end{figure}

\cref{fig:velocity_time} depicts the linear velocities in x- and y directions, as well as the rotational speed on the yaw axis commands sent to the avatar robot during the Day~2 run. Task~4, 6, 7, and 9 required manly manipulation capabilities, which asked the operator to position the avatar robot precisely. \cref{fig:velocity_time} shows that only minor corrections were made after reaching the manipulation tasks (small forward movement at the beginning of Task~6 and 9, etc.). Especially, only corrections in one direction (e.g. only forward, or only right) were made, which proofs the controller to be easy and intuitive to use since the operator never overshoot the target.
The majority of locomotion commands were intended, and the design was able to allow intuitive locomotion of the system.

\section{Conclusion}

In this paper, we have presented a novel locomotion controller for self-centering hands-free operation within the context of an immersive avatar robot. 
The controller has demonstrated its proficiency in the ANA Avatar XPRIZE competition, wherein the NimbRo team emerged victorious. The controller showed to be intuitive throughout our experiments and allowed for locomotion control in various tasks like traversing the robot with and without obstacles, and aligning in case of manipulation and interaction tasks.
The system description and 3D models enable telepresence and VR researchers to replicate the controller.
Future research directions could include integrating vibrational or haptic feedback into the locomotion controller.

\printbibliography

\end{document}